# Clustering Prominent Named Entities in Topic-Specific Text Corpora

*Completed Research Full Papers*

**Abdulkareem Alsudais**
College of Computer Engineering and Sciences
Prince Sattam Bin Abdulaziz University
A.Alsudais@psau.edu

**Hovig Tchalian**
Drucker School of Management
Claremont Graduate University
Hovig.Tchalian@cgu.edu

## Abstract

Named Entity Recognition (NER) refers to the computational task of identifying real-world entities in text documents. A research challenge is to use computational techniques to identify and utilize these entities to improve several NLP applications. In this paper, a method that clusters prominent names of people and organizations based on their semantic similarity in a text corpus is proposed. The method relies on common named entity recognition techniques and word embeddings models. Semantic similarity scores generated using word embeddings models for named entities are used to cluster similar entities of the people and organizations types. A human judge evaluated ten variations of the method after it was run on a corpus that consists of 4,821 articles on a specific topic. The performance of the method was measured using three quantitative measures. The results of these three metrics demonstrate that the method is effective in clustering semantically similar named entities.

## Introduction

Researchers and scientists often turn to textual data to study social phenomena (Baker et al. 2013; Dimaggio et al. 2013; Jaworska and Krishnamurthy 2012). Mining those data allows scholars to generate insights about the grammatical features, linguistic content and social context of such phenomena (Li and Dash 2010; Yuan et al. 2016). Researchers use common text mining methods such as topic modeling and entity extraction to explore text corpora, identifying patterns, generating insights, and confirming and disconfirming hypotheses. Generating genuine insights through these methods requires the development of novel algorithms but, equally, their rigorous application in methods that are being continuously developed. Therefore, novel and data science-based text mining solutions that produce specific, consistent and relevant results are especially beneficial for several relevant systems and tools (Liu et al. 2017).

When examining a text corpus, one useful step in generating specific, relevant insights is to identify prominent entities such as individuals and organizations presents in the corpus. Named Entity Recognition (NER) refers to the computational task of identifying real-world entities in text documents. Once identified and tagged, named entities may be utilized in information systems to enhance applications and tools as well as to generate useful information.

Word embeddings refer to models that create vector representations for words or phrases in a text corpus. Successful models for generating vector representations for words and phrases such as Skip-gram and CBOW (Mikolov et al. 2013), GloVe (Pennington et al. 2014), and others (Levy and Goldberg 2014; Turian et al. 2010) have proven successful in performing various language-related analytics tasks. One of the most beneficial features of these new word embedding methods is a function that captures semantic similarities across words and phrases in a corpus. This similarity feature may be used, for example, to capture and detect the semantic changes in the meaning of particular words over a pre-defined period of time (Hamilton et al. 2016; Kulkarni et al. 2015). This application demonstrates the potential of using this feature to solve new and challenging questions in the text-mining domain. The objective of this paper is to resolve one of these questions: Is it possible to create an NLP pipeline that reliably and consistently clusters named entities (people and organizations) in a corpus by utilizing their semantic similarity as generated by a semantic word embeddings model?

In this paper, semantic similarities generated by Skip-gram and CBOW are employed to cluster the most prominent named entities in a corpus. Clustering named entities is a recognized approach for leveraging tagged entities (Hasegawa et al. 2004). When clustering named entities, similar named entities are grouped within the same cluster based on their similarity values, which are detected using a purpose-built, reliable similarity measure. Thus, an underlying metric for capturing the similarities between the entities must be utilized. The method proposed in this paper offers a rigorous and reliable alternative to simple clustering methods, offering information systems researchers a replicable method for generating results and insights useful for various text mining and NER applications. When training a Skip-gram or CBOW model on a topic-specific corpus, one may discover relationships between words and entities in the collection that are otherwise latent and difficult to detect. According to Thomas Mikolov, the chief scientist behind the two models, the learned word vectors are only effective if they are used to accomplish other tasks. For example, Mikolov suggests using word vectors and K-means clustering to create classes for the words in the collection. While that is useful and currently achievable using the word2vec software, additional modifications and customizations are required to implement problem-specific solutions that build on that idea. Generating additional insights in a consistent, replicable way from the relationship among tagged named entities therefore offers researchers numerous benefits including enhancing Question Answering (QA) systems and automating the population process in ontologies (Marrero et al. 2013).

This paper develops a novel, reliable and replicable method for semantically clustering similar named entities, primarily of the "people" and "organization" type, based on the similarity scores as generated by a Skip-gram or CBOW model. It does so by: (1) using NER to locate named entities in the text, such as the names of real-world organizations; (2) using semantic similarity vectors as captured by a word embedding model; (3) deploying a reliable algorithm to cluster the entities according to the resulting vectors. The preliminary empirical findings and quantitative evaluation of the method demonstrate that the it is successful in clustering and grouping named entities that have similar roles in the text or are members of the same abstract classes.

## Method and Data

In this section, the method proposed to identify, and cluster named entities in a text corpus is described. The method relies on three major components: named entity recognition, word embeddings, and clustering. Fig. 1 demonstrates the three major steps in the method and the relevant tasks that take place in each step. In the first step, named entities in the corpus are identified and then ranked according to the frequency in which they were used in the articles of the corpus. In the second step, a word embeddings model is applied on the collection to generate semantic similarity scores across the named entities in the corpus. In the final step, a clustering operation is performed on a symmetrical matrix that consists of top named entities to identify clusters of similar individuals and organizations in the corpus. This section describes the three steps, the rationale behind each task proposed for each step, and the underlying assumptions that were made.

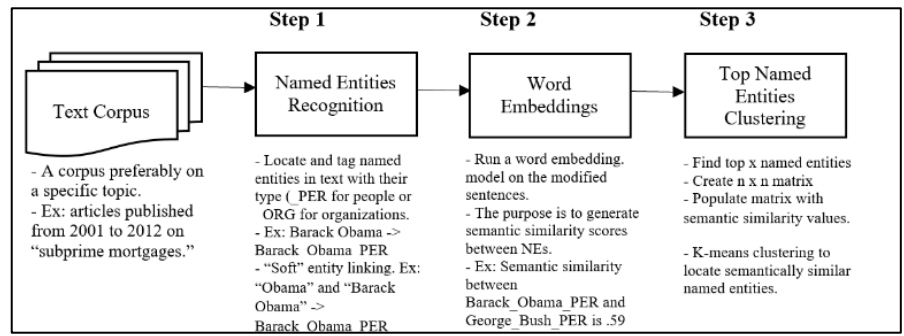

**Figure 1. Summary of Phases in the Method**

### *Named Entity Recognition*

Named Entity Recognition (NER) is the task of processing a body of text and classifying segments in the text that are named entities. Several highly accurate and reliable algorithms for capturing named entities currently exist (Finkel et al. 2005; Mccallum and Li 2003). Recent work in the area has focuses on increasing the accuracy of named entity taggers by leveraging advances in neural networks (Chiu and Nichols 2016; Lample et al. 2016). However, these new models are only slightly more effective than their traditional counterparts. Several named entity extraction methods target unique types of corpora such as tweets (Habib and Van Keulen 2015) and biomedical data (Habibi et al. 2017) and certain methods are designed for specific languages such as Chinese (Peng and Dredze 2016) and Arabic (Oudah and Shaalan 2016).

There are multiple categories for named entities including the names of people, organizations, locations, and time units. Since the primary motivation of the proposed method is to semantically cluster similar people and organizations, named entities of other types such as "location" and "time" have been discarded. However, other named entity types should be considered in future extensions of the method.

In this paper, Stanford NER (Finkel et al. 2005) is used in the first step in the proposed method. Stanford NER, which is based on an older mechanism of detecting named entities, is a highly accurate tagger that performs well on text extracted from newspaper articles. According to (Finkel et al. 2005), Stanford NER achieved an F1 score (a composite measure employed to determine the accuracy of a classifier) of 86% when trained on the CoNLL-2003 named entity dataset (Tjong et al. 2003). The NER process followed in this paper is initiated by tokenizing each article in the collection individually, in order to locate the set of sentences in the article. Subsequently, the Stanford NER program is run on each extracted sentence.

The next task in this step is entity linking. Entity linking is the process of matching different segments of text that belong to the same real-world entity. For instance, "Barack Obama" and "Barack H. Obama" refer to the same person, and an entity linking procedure should be able to recognize this and match the two variations of the name accordingly. To increase the performance of the proposed method, a "soft" entity linking process is performed on tagged names of people at the article-level. For organizational names, no attempts were made to match entities. For example, "SEC" and "Securities and Exchange Commission" were captured as separate entities. This is a limitation of the current method that will be explored in future extensions.

The final task in the named entities recognition process involves replacing named entities in each article with their full names as captured in the "entity linking" task with their corresponding entity type. For example, each occurrence of the names "Barack Obama" and "Barack H. Obama" would be resolved to "Barack_Obama_PER" in a given article. The underscores replace empty spaces so that the word embeddings model can recognize the name as a single entity and then process it accordingly. The tag "_PER" is used to indicate that the named entity is of the "person" type while the tag "_ORG" is used to indicate that the named entity is of the "organization" type. The types are added to the names so that they may be later retrieved using a word embeddings model. In future extensions of this work, reliable coreference resolution methods will be used to accomplish this task.

### *Word Embeddings*

Word embeddings models are better at capturing semantic similarities between words as compared to traditional "counting" methods that use either Positive Pointwise Mutual Information (PPMI) or Local Mutual Information (LMI) to weight the features in the vectors (Baroni et al. 2014). Researchers have demonstrated how this similarity feature available in word embeddings models may be employed to accomplish several challenging language-related tasks.

Once all the named entities in the corpus are tagged, the next step is to run a word embeddings model on the modified corpus. The objective of this step is to detect the semantic similarity across named entities in the corpus. In this study, two popular word embeddings models, CBOW and Skip-gram, are tested and applied on a text corpus. Both models are used to investigate whether the performance of the proposed method is affected by the model that is employed to capture the similarity between the named entities. The implementation of CBOW and Skip-gram in the python package gensim (Rehurek and Sojka 2010) is used to run the models. Running the two models generates word vectors for named entities in the collection. These vectors and similarity scores are used in the following step to cluster the named entities.

### *Clustering Top Named Entities*

Clustering is a data mining technique that is used to group similar items based on an underlying similarity metric. Clustering may be applied to different types of data including text. In the text mining domain, clustering may be used to group text segments such as words, terms, sentences, topics, or documents. The results of clustered text data may be utilized to enhance text mining tasks such as corpus summarization and document classification (Aggarwal and Zhai 2012). In this paper, clustering is applied on a matrix that consists of the top named entities in a collection and their similarity values. The purpose of this clustering procedure is to discover semantic roles, labels, and categories.

Several clustering algorithms currently exist, and for this paper, K-means is used as the clustering algorithm. In K-means, objects are segmented into different groups where each cluster contains at least one item. No items may be placed into more than one cluster. The number of clusters 'K' must be specified prior to initiating the algorithm. Choosing the most appropriate K is an ongoing research challenge. For the proposed method, K is calculated by dividing the number of top terms to be clustered by the number 10, a common practice when using K-means clustering. Implementing a different method of calculating K or using a clustering algorithm other than K-means such as the silhouette method or X-means clustering might produce better results. These methods were not employed in this study and present opportunities for future research and refinement.

The final task in the method is to identify and cluster the N top named entities in the corpus. To accomplish this goal, a symmetrical N x N matrix is first constructed. In this matrix, the columns and rows consist of all the named entities in the top N named entities list. For each entity in the rows section, a vector is created. The vector consists of the similarity values as generated by the word embeddings model between the entity and the corresponding entities in the columns section. The concluding step is to then cluster the entities in the rows section using K-means clustering.

The top terms lists are generated by counting the number of articles for each named entity in the corpus where the named entity appeared. Accordingly, the N named entities with the highest numbers are selected. The number of articles was used as a unit of measure rather than the raw frequency to provide additional rigor and to avoid assigning a large weight to named entities that appeared several times in the corpus but in only a small number of articles. Using a larger corpus is recommended, since a larger set of terms is more likely to accurately and precisely identify a subset of top terms. To maintain the automation of the method, qualitatively refining and filtering terms in the top terms lists was avoided. Thus, terms that are erroneously tagged as named entities and appear in the top terms list are kept, even though they contribute negatively to the results. The rationale behind this approach is to refrain from using experts' opinions to modify the lists and to maintain a fully automated process. The approach allows greater replicability of the method by relying exclusively on quantitative results, without subjective modifications by experts, while sacrificing very little in accuracy and precision.

In this paper, five different top named entities lists are tested. The lists differ in two core variables: (1) the number of terms in the list (either 100 or 200) and (2) the type of named entities in the list. Two of the lists contains entities of the "people" type, two of the lists contains entities of the "organizations" type, and one list contains entities of both the "people" and "organization" types.

### *Data*

The method proposed in this paper was developed for use on a large text corpus. A large corpus is required because word embeddings models perform best on large corpora and often fail when applied on small ones. In this study, a large text corpus on "corporate governance" was used (Alsudais and Tchalian 2016). The corpus consists of 4,821 articles published on the topic between 1975 and 2004.

# Experiment

In this section, an experiment to test and evaluate the proposed method is explained. The method was tested on the "corporate governance" corpus described in the previous section. Five top terms lists were investigated. The lists were: (1) top 100 tagged entities of the "person" type (T100_P); (2) top 200 tagged entities of the "person" type (T200_P); (3) top 100 tagged entities of the "organization" type (T100_O); (4) top 200 tagged entities of the "organization" type (T200_O); and (5) a combined list of the top 100 tagged entities of both the "person" and "organization" types (T200_PO). Both CBOW and Skip-gram models were used to generate vectors for the named entities. For both models, the number of dimensions was set to 500 and the window size was set to 10. For each one of the five lists, two N x N matrices were constructed with the semantic similarity between the variables in the row and column as the values of the cells. Therefore, ten matrices were constructed. The values in the first matrix were derived from the results generated using CBOW whereas the values in the second were derived from the results generated using Skip-gram. The purpose of testing these variations is to investigate whether the changes in the models or the type of input would generate improved results.

## *Evaluation Metrics*

For quantitative evaluation, three metrics were used. The evaluation process relied in using a domain expert with a Ph.D. in Economics who manually reviewed and evaluated the resulting clusters. The expert was asked to complete two tasks: (1) indicate whether a logical and semantic role or class might be inferred and used to label at least the majority of terms for each cluster identified by the method; and, if that was the case, (2) indicate for each named entity in the cluster whether the entity belong to the category identified for the cluster. Accordingly, three evaluation metrics were used. In text mining, these metrics are typically used to evaluate topic modeling and clustering algorithms. The first metric is a coherence measure that is used to evaluate the average coherence of the clusters. The second metric is a standard precision measure that is employed to quantify how "precise" the method is in assigning terms to clusters. The third metric is a coherent clusters metric that is used to quantify the number of clusters that comprise a meaningful class or category and may be tagged with a class label.

The coherence measure is a metric that is commonly used when evaluating topic models and it has been used in other studies (Qiang et al. 2016; Xie and Xing 2013; Xie et al. 2015). There are many automated methods for generating coherence scores. However, it has been argued that coherence values that rely on human judgement are still superior (Röder et al. 2015). Additionally, many of these automated metrics rely on co-occurrence values between two terms at the sentence or article levels. This makes the metrics unsuitable for use in this work since two terms might be accurately placed in the same cluster according to the proposed method even if their co-occurrences value was low. The terms would be placed in the same cluster since they are semantically similar because they share similar neighboring words. An example of this is a cluster labeled "corrupt CEOs" where accurately placed terms in the cluster would be names of corrupt CEOs who may have similar neighboring words such as "CEO," "crooked," and "corrupt" yet still have low values as generated by traditional co-occurrence-based metrics.

A human judge is therefore used to evaluate the coherence or clusters for purposes of developing and testing the method proposed in this paper. The coherence of a cluster is calculated based on the ratio of the number of relevant terms in the cluster to the total number of terms in the cluster. The judge is asked to infer a label for an examined cluster. If a label can be found, the judge manually tags irrelevant terms in the cluster that do not fit and therefore do not belong to the cluster. Similar to the evaluation process in (Qiang et al. 2016), all terms within a cluster are considered irrelevant if the judge is unable to infer a semantic label or class for the cluster. The overall coherence for a method is calculated by averaging the individual coherence values for the clusters in the set. The formula for calculating the coherence value of a cluster is:

$$\frac{\textit{Number of relevant named entites in cluster}}{\textit{Number of named entities in the cluster}}$$

For the precision measure, the values are calculated by considering the number of True Positive (TP) and False Positive (FP) named entities in the results previously tagged by the judge. The standard formula for calculating the precision value is used. A TP named entity is defined as a named entity that has been accurately placed in a cluster and confirmed by a judge. Conversely, a FP named entity is an entity the judge has tagged as an entity that has been inaccurately placed in the cluster. Similar to the process followed for the coherence measure, all entities in a cluster are tagged as false positive (Type I error) if the judge is unable to infer a semantic label or class for the cluster.

The third measure used in this paper is a coherent clusters measure. The purpose of this measure is to quantify the number of clusters that are labeled as valid by the judge. The formula for calculating the coherent clusters measure is:

$$\frac{\textit{Number of clusters tagged as accurate}}{\textit{Total number of clusters K}}$$

The judge was instructed to tag a cluster as accurate if at least the majority of the entities in the cluster could be grouped under the same semantic class or label. An example of a label for a cluster would be "corrupt CEOs" or "companies that faced scandals in the past." Since the previous two measures were heavily influenced by false positive terms, this metric could be used to provide additional insights on the performance of the method and to measure the method's ability to produce strong clusters even when some of the clusters contain several noisy terms. Furthermore, this measure might be employed to hint at how the results of the method might be interpreted by a domain expert who might manually eliminate problematic terms that negatively affect otherwise strong clusters.

## Results

| List: | | T100_P | T200_P | T100_O | T200_O | T100_PO | Average |
|---|---|---|---|---|---|---|---|
| K size: | | 10 | 20 | 10 | 20 | 20 | |
| Coherence measure | CBOW | 72.5% | 64.8% | 73.4% | 73.9% | 74.0% | 71.7% |
| | SkipGram | 81.7% | 61.5% | 79.6% | 76.6% | 77.6% | 75.4% |
| Precision measure | CBOW | 86.5% | 70.0% | 92.7% | 71.5% | 83.8% | 80.9% |
| | SkipGram | 90.8% | 67.0% | 78.0% | 86.0% | 86.0% | 81.6% |
| Coherent clusters measure | CBOW | 90.0% | 75.0% | 90.0% | 80.0% | 90.0% | 85.0% |
| | SkipGram | 90.0% | 75.0% | 80.0% | 90.0% | 90.0% | 85.0% |

**Table 1. Summary of Results**

In this section, results of running the method on the "corporate governance" corpus using the previously described top named entities lists as well as both CBOW and Skip-gram are described. The judge was asked to analyze the generated clusters using ten variations of the method. The judge was given the results of the ten variations without any indicators of the underlying model or list used.

The method was quantitatively evaluated based on the three quantitative measures: coherence measure, precision measure, and coherent clusters measure. The results for the three measures indicate that the method was successful in detecting semantic categories of similar people and organizations. With respect to the performance of CBOW and Skip-gram, the results were slightly more accurate when Skip-gram. These findings are identical to those of previous studies that compared the performance of these two methods for various language-related tasks (Chen et al. 2015). While improvements can be made, these results demonstrate the method's ability to detect semantically similar clusters of people and organizations in a text corpus. The quantitative results suggest that the method performed best when the lists included only people or only organizations. The best results were achieved when the top 100 named entities tagged with the "people" type was processed using Skip-gram and then clustered using a K size of 10. Table 1 includes the detailed results of the three measures based on the evaluation performed by the judge.

For the coherence measure, the best result was achieved when the list of top 100 people was used with Skip-gram. Results when Skip-gram was used were better. Results of the precision measure were similar to those of the coherence measure, with Skip-gram averaging a few percentage points higher than CBOW as well. The best overall precision value was attained when CBOW and the T100_O list were used. The lowest value was 67% suggesting that the method provided useful information even when performing at its worst. Several inaccurately tagged named entities that were meaningless and do not represent a real-world person

*Clustering Prominent Named Entities in Topic-Specific Text Corpora*or organization, such as "John" and "Messes," negatively affected the performance of the precision measure as they were flagged as unfit by the judge.

The third and final measure was the coherent clusters measure, which was based on counting the number of clusters that the judge deemed as being logical and accurately representing a semantic category. The results of this measure were the most promising. For many of the variations, the judge found that 80% of the groups captured by the method were valid and coherent. This suggests that the method may be produce more accurate results and demonstrate greater precision if a domain expert examines the results and removes unfit terms from each cluster. Users of the method can use these tradeoffs to help balance automation with error rates for individual cases.

Selected clusters discovered using variation of the method are displayed in Table 2. In the table, named entities the judge deemed unfit for a discovered cluster are designated in italics. The second line includes the labels as inferred by the judge for the cluster. The third row contains the terms within the clusters. The terms in italic are ones that the judge declared they have been inaccurately placed in this cluster.

| Method: | T100_O & skip-gram | | T100_P & skip-gram | | T100_O & CBOW | |
|---|---|---|---|---|---|---|
| Cluster (labeled by judge): | Companies experiencing financial scandals in early 2000s and their auditors | Investment Banks | Company leaders involved in financial or investment scandal | Corporate Governance thought leaders | Regulatory bodies | Universities |
| Entities in the cluster: | 1. Enron Corp<br>2. Enron<br>3. WorldCom<br>4. WorldCom Inc.<br>5. Tyco International Ltd.<br>6. Pricewaterhouse Coopers<br>7. MCI<br>8. Tyco International<br>9. Tyco | 1. Merrill Lynch<br>2. Goldman Sachs<br>3. Deutsche Bank<br>4. Morgan Stanley<br>5. Credit Suisse First Boston<br>6. Lehman Brothers | 1. Arthur Andersen<br>2. Dennis Kozlowski<br>3. Bernard Ebbers<br>4. Kenneth Lay<br>5. *Morgan Stanley*<br>6. *Fannie Mae*<br>7. Martha Stewart<br>8. Mark Swartz<br>9. *Freddie Mac* | 1. Ira Millstein<br>2. John Coffee<br>3. *Weil*<br>4. Graef Crystal<br>5. Joseph Grundfest<br>6. *Spencer Stuart*<br>7. Jay Lorsch | 1. Congress<br>2. Senate<br>3. Federal Reserve<br>4. House<br>5. Delaware Chancery Court<br>6. Senate Banking Committee<br>7. European Union<br>8. Supreme Court | 1. Columbia University<br>2. Harvard Business School<br>3. *Gotshal & Manges*<br>4. Stanford University<br>5. Harvard University<br>6. Harvard |

**Table 2. Examples of Discovered Clusters**

# Conclusion

In this paper, a method that utilizes existing and accepted techniques to cluster named entities of the "person" and "organization" types in a topic-specific corpus based on the semantic similarities between the entities as found by the CBOW and Skip-gram models is introduced. The method was tested on a corpus that consists of news articles containing the term "corporate governance" published in four of the leading newspapers in the United States between 1978 and 2004. A domain expert reviewed the results. Three quantitative metrics were used to quantify the performance of the method. The results demonstrate the effectiveness of the method when evaluated using these quantitative metrics.

Observations with respect to the results of the process performed by the judge suggest that the method effectively captures semantic clusters of named entities. For example, when using Skip-gram and the list of top 100 organizations, investment banks was one of the identified clusters. The investment banks cluster included banks such as "Morgan Stanley" and "Merrill Lynch". Another cluster found while using Skip-gram and the list of top 100 organizations was a narrower cluster that included companies dealing with financial crises in the early 2000s and two of their auditors. This kind of information might be valuable for non-domain experts as it captures information that is specific to a topic-specific corpus. An additional observation was that changing the list of top named entities affected the performance of clustering. Using lists of only people or only organizations generated more coherent results. Furthermore, results indicated that using Skip-gram generated more coherent clusters when compared to CBOW. The result of the quantitative measures agreed with these observations.

Another observation is that some of the clusters were narrow and defined whereas others were broader and more abstract. Narrow clusters are ones that are unique to the dataset and the analyzed corpus. Example of such clusters are "corporate governance experts" and "companies in crisis in the 1980s." On the other hand, broad clusters are those that represent semantic categories that are more universal such as "universities" and "financial services firms." Table 3 includes a sample of narrow and broad clusters.

| Narrow Clusters | Broad Clusters |
|---|---|
| Corporate governance thought leaders | Universities |
| Companies in crisis in 1980s | Company CEOs and founders |
| Companies experiencing financial scandals in early 2000s | Large, publicly-traded companies |
| Corporate governance experts | Financial services firms |
| Company executives involved in financial or investment scandal | Company executives |
| Corporate governance associations | American business executives |
| Corporate governance experts | American government executives |
| SEC leaders (chairs or commissioners) | |

**Table 3. Examples of narrow and broad clusters**

While the results are promising, there remain several limitations in the method and the paper. First, the method was tested and evaluated on a single corpus. Thus, testing the method on a different dataset might produce different results. Additionally, latent issues in the method might be revealed. Second, due to the historical context of the corpus used in this paper, it was difficult for the judge to fairly evaluate some of the extracted clusters. Some of the clusters represented narrow and specific categories that might be challenging to interpret and label, even for domain experts, suggesting that some of the reported results may have fewer errors than those reported in the paper. Finally, while entity linking techniques were used for entities of the person type, similar techniques were not employed for entities of the organization type. Thus, many replicates existed in the top organizations lists.

There are also several areas where the method may be enhanced in future work. First, the techniques used to generate the top terms lists may be improved. Currently, the top named entities lists are generated by counting the number of articles that named entities appear within, and then ranking the named entities accordingly. It is possible that the use of more complex alternatives that leverage additional information pertaining to the entities could significantly improve the results. Second, the current method assumes a fully automated process with no expert intervention. The coherence of the clusters will improve if a domain expert provides expert feedback at various points. This expert feedback may occur after generating the top terms lists by having the expert flag terms that are inaccurately tagged as named entities.